\documentclass[11pt,a4paper]{article}
\usepackage[hyperref]{emnlp2020}
\usepackage{times}
\usepackage{latexsym}

\usepackage{algpseudocode,algorithm,algorithmicx}

\usepackage{times}
\usepackage{latexsym}

\usepackage{microtype}

\newcommand\tightdots{\hbox to 1em{.\hss.\hss.}}

\newcommand{\commaNum}[1]{\num[group-separator={,}]{#1}}

\newcommand{\importanceRankingName}{word importance ranking }
\newcommand{\importanceRankingNameCap}{Word Importance Ranking }
\newcommand{\importanceRankingNameAbbrev}{WIR }

\usepackage{enumitem}
\setlist[itemize]{leftmargin=*}

\usepackage{amsmath, amsthm, amssymb}
\usepackage{microtype}
\usepackage{graphicx}
\usepackage{subfigure}
\usepackage{booktabs} %
\usepackage{times}
\usepackage{latexsym}
\usepackage{multirow}
\usepackage{color}
\usepackage{siunitx}

\graphicspath{{figures/}{../figures/}}

\usepackage{pifont}%
\usepackage{amssymb}%
\usepackage{pifont}%
\newcommand{\cmark}{\ding{51}}%
\newcommand{\xmark}{\ding{55}}%

\usepackage{hyperref}

\DeclareMathOperator*{\argmax}{arg\,max}

\include{macros}
\def\x{{\mathbf x}}

\usepackage{placeins}

\usepackage{lipsum}
\usepackage{titlesec}

\titlespacing\section{0pt}{5pt plus 1pt minus 1pt}{2pt plus 1pt minus 1pt}
\titlespacing\subsection{0pt}{1pt plus 0pt minus 1pt}{1pt plus 0pt minus 0pt}
\titlespacing\subsubsection{0pt}{1pt plus 0pt minus 1pt}{0pt plus 0pt minus 0pt}
\titlespacing{\paragraph}{2pt}{2pt}{0pt}[0pt]  
\usepackage{etoolbox}
\makeatletter
\preto{\@tabular}{\parskip=5pt}
\makeatother

\usepackage{enumitem}
\setlist[itemize]{leftmargin=*}
\setlist{nosep}

\usepackage[utf8]{inputenc} %

\allowdisplaybreaks

\usepackage{flushend}

\usepackage{microtype}

\aclfinalcopy 

\title{\textit{Searching for a Search Method}: Benchmarking Search Algorithms for Generating NLP Adversarial Examples}

\author{Jin Yong Yoo\thanks{\textsuperscript{*} Equal contribution. Code implementation shared via  \url{https://github.com/QData/TextAttack-Search-Benchmark}.}, ~
John X. Morris\footnotemark[1],
~ Eli Lifland\, 
~ Yanjun Qi \\
Department of Computer Science, University of Virginia \\
\{\href{mailto:jy2ma@virginia.edu}{jy2ma}, \href{mailto:yq2h@virginia.edu}{yq2h}\}@virginia.edu
\\
}

\date{}

\begin{document}
\maketitle

\begin{abstract}
We study the behavior of several black-box search algorithms used for generating adversarial examples for natural language processing (NLP) tasks. We perform a fine-grained analysis of three elements relevant to search: search algorithm, search space, and search budget. When new search algorithms are proposed in past work, the attack search space is often modified alongside the search algorithm. Without ablation studies benchmarking the search algorithm change with the search space held constant, one cannot tell if an increase in attack success rate is a result of an improved search algorithm or a less restrictive search space. Additionally, many previous studies fail to properly consider the search algorithms' run-time cost, which is essential for downstream tasks like adversarial training. Our experiments provide a reproducible benchmark of search algorithms across a variety of search spaces and query budgets to guide future research in adversarial NLP. Based on our experiments, we recommend greedy attacks with word importance ranking when under a time constraint or attacking long inputs, and either beam search or particle swarm optimization otherwise.

\end{abstract}
\setlength{\parskip}{0.35em}

\section{Introduction}

Research has shown that current deep neural network models lack the ability to make correct predictions on adversarial examples \cite{szegedy2013intriguing}. The field of investigating the adversarial robustness of NLP models has seen growing interest, both in contributing new attack methods \footnote{In this work, we use ``adversarial example generation methods'' and ``adversarial attacks'' interchangeably.} for generating adversarial examples \cite{Ebrahimi2017HotFlipWA, Gao2018BlackBoxGO, alzantot2018generating, Jin2019TextFooler, pwws-ren-etal-2019-generating, pso-zang-etal-2020-word} and better training strategies to make models resistant to adversaries \cite{jia2019certified,goodfellow2014explaining}.

Recent studies formulate NLP adversarial attacks as a combinatorial search task and feature the specific search algorithm they use as the key contribution \cite{Zhang19Survey}. The search algorithm aims to perturb a text input with language transformations such as misspellings or synonym substitutions in order to fool a target NLP model when the perturbation adheres to some linguistic constraints (e.g., edit distance, grammar constraint, semantic similarity constraint) \cite{Morris2020TextAttackAF}. Many search algorithms have been proposed for this process, including varieties of greedy search, beam search, and population-based search.

The literature includes a mixture of incomparable and unclear results when comparing search strategies since studies often fail to consider the other two necessary primitives in the search process: the search space (choice of transformation and constraints) and the search budget (in queries to the victim model). The lack of a consistent benchmark on search algorithms has hindered the use of adversarial examples to understand and to improve NLP models. In this work, we attempt to clear the air by answering the following question: \emph{Which search algorithm should NLP researchers pick for generating NLP adversarial examples?}

We focus on black-box search algorithms due to their practicality and prevalence in the NLP attack literature. Our goal is to understand to what extent the choice of search algorithms matter in generating text adversarial examples and how different search algorithms compare when we hold the search space constant or when we standardize the search cost. We select three families of search algorithms proposed from literature and benchmark their performance on generating adversarial examples for sentiment classification and textual entailment tasks. Our main findings can be summarized as the following:
\begin{itemize}[noitemsep,topsep=0pt]
    \item Across three datasets and three search spaces, we found that beam search and particle swarm optimization are the best algorithms in terms of attack success rate.
    \item When under a time constraint or when the input text is long, greedy with \importanceRankingName is preferred and offers sufficient performance.
    \item Complex algorithms such as \texttt{PWWS} \cite{pwws-ren-etal-2019-generating} and genetic algorithm \cite{alzantot2018generating} are often less performant than simple greedy methods both in terms of attack success rate and speed.
\end{itemize}

\section{Background}

\subsection{Components of an NLP Attack}
\label{method:basics}

\citet{morris2020reevaluating} formulated the process of generating natural language adversarial examples as a system of four components: a goal function, a set of constraints, a transformation, and a search algorithm.

Such a system searches for a perturbation from $\x$ to $\x'$ that fools a predictive NLP model by both achieving some goal (like fooling the model into predicting the wrong classification label) and fulfilling certain constraints. The search algorithm attempts to find a sequence of transformations that results in a successful perturbation.

\subsection{Elements of a Search Process}

\paragraph{Search Algorithm:}~Recent methods proposed for generating adversarial examples in NLP  frame their approach as a combinatorial search problem. This is necessary because of the exponential nature of the search space. Consider the search space for an adversarial attack that replaces words with synonyms: If a given sequence of text consists of $W$ words, and each word has $T$ potential substitutions, the total number of perturbed inputs to consider is $(T + 1)^W - 1$. Thus, the graph of all potential adversarial examples for a given input is far too large for an exhaustive search.

While heuristic search algorithms cannot guarantee an optimal solution, they can be employed to efficiently search this space for a valid adversarial example. Studies on NLP attacks have explored various heuristic search algorithms, including beam search \cite{Ebrahimi2017HotFlipWA}, genetic algorithm \cite{alzantot2018generating}, and greedy method with \importanceRankingName \cite{ Gao2018BlackBoxGO,Jin2019TextFooler, pwws-ren-etal-2019-generating}.

\paragraph{Search Space:}~In addition to its search method, an NLP attack is defined by how it chooses its search space. The search space is mainly determined by two things: a transformation, which defines how the original text is perturbed (e.g. word substitution, word deletion) and the set of linguistic constraints (e.g minimum semantic similarity, correct grammar) enforced to ensure that the perturbed text is a valid adversarial example. A larger search space corresponds to a looser definition of a valid adversarial example. With a looser definition, the search space includes more candidate adversarial examples. The more candidates there are, the more likely the search is to find an example that fools the victim model – thereby achieving a higher attack success rate \cite{morris2020reevaluating}.

\paragraph{Search Cost/Budget:}~Furthermore, most works do not consider the runtime of the search algorithms. This has created a large, previously unspoken disparity in runtimes of proposed works. Population-based algorithms like \citet{alzantot2018generating} and \citet{pso-zang-etal-2020-word} are significantly more expensive than greedy algorithms like \citet{Jin2019TextFooler} and \citet{pwws-ren-etal-2019-generating}. Additionally, greedy algorithms with \importanceRankingName are linear with respect to input length, while beam search algorithms are quadratic. In tasks such as adversarial training, adversarial examples must be generated quickly, and a more efficient algorithm may preferable– even at the expense of a lower attack success rate.

\subsection{Evaluating Novel Search Algorithms}

Past studies on NLP attacks that propose new search algorithms often also propose a slightly altered search space, by proposing either new transformations or new constraints. When new search algorithms are benchmarked in a new search space, they cannot be easily compared with search algorithms from other attacks.

To show improvements over a search method from previous work, a new search method must be benchmarked in the search space of the original method. However, many works fail to set the search space to be consistent when comparing their method to baseline methods. For example, \citet{Jin2019TextFooler} compares its \texttt{TextFooler} method against \citet{alzantot2018generating}'s method without accounting for the fact that \texttt{TextFooler} uses the Universal Sentence Encoder \cite{Cer18USE} to filter perturbed text while \citet{alzantot2018generating} uses Google 1 billion words language model \cite{google-1-billion-2013}. A more severe case is \citet{zhang2019mha}\footnote{\citet{zhang2019mha} is not considered in this paper due to failure to replicate its results.}, which claims that its Metropolis-Hastings sampling method is superior to \citet{alzantot2018generating} without setting any constraints – like \citet{alzantot2018generating} does – that ensure that the perturbed text preserves the original semantics of the text. 
 
We do note that \citet{pwws-ren-etal-2019-generating} and \citet{pso-zang-etal-2020-word} do provide comparisons where the search spaces are consistent. However, these works consider a small number of search algorithms as baseline methods, and fail to provide a comprehensive comparison of methods proposed in the literature.

\section{Benchmarking Setup}

\subsection{Defining Search Spaces}
As defined in Section~\ref{method:basics}, each NLP adversarial attack includes four components: a goal function, constraints, a transformation, and a search algorithm. We define the \emph{attack search space} as the set of perturbed text $\x'$ that are generated for an original input $\x$ via valid transformations and satisfy a set of linguistic constraints. The goal of a search algorithm is to find those $\x'$ that achieves the attack goal function (i.e. fooling a victim model)
as fast as it can.

\paragraph{Word-swap transformations:}~ Assuming $\x=(x_1,\dots, x_i,\dots, x_n)$, a perturbed text $\x'$ can be generated by  swapping $\x_i$ with altered $\x'_i$. The swap can occur at word, character, or sentence level, depending on the granularity of $\x_i$. Most works in literature choose to swap out words; therefore, we choose to focus on word-swap transformations for our experiments.

\paragraph{Constraints:}~ \citet{morris2020reevaluating} proposed a set of linguistic constraints to enforce that $\x$ and perturbed $\x'$ should be similar in both meaning and fluency to make $\x'$ a valid \textit{potential adversarial} example. This indicates that the search space should ensure $\x$ and $\x'$ are close in semantic embedding space. Multiple automatic constraint ensuring strategies have been proposed in the literature. For example, when swapping word $\x_i$ with $\x'_i$, we can require that the cosine similarity between word embedding vectors $e_{\x_i}$ and $e_{\x'_i}$ meet certain minimum threshold. More details on the specific constraints we use are in Section~\ref{exp:space}.

Now we use notation $T(\x) = \x'$ to denote transformations perturbing $\x$ to $\x'$, and assume the $j-th$ constraints as Boolean functions $C_j(\x, \x')$ indicating whether $\x'$ satisfies the constraint $C_j$. Then, we can define the search space $S$ mathematically as:
\begin{equation}
    S(\x)=\{T(\x)|C_j(\x, T(\x))\; \forall j\in[m] \}
\end{equation}
The goal of a search algorithm is to find $\x'\in S(x)$ such that $\x'$ succeeds in fooling the victim model. 
Table \ref{table:search-spaces} describes three search spaces we use to benchmark the search algorithms. Details of transformations and constraints used in defining these search spaces are in Appendix Section~\ref{exp:space}. 
\begin{table}[tbh]
\centering
\scalebox{0.85}{
\begin{tabular}{|c|p{2.5cm}|p{5cm}|}
\hline
  & \textbf{Transformation} & \textbf{Constraints} \\ \hline
1 & Counter-fitted GLOVE Word Embedding & Word embedding similarity, BERTScore, POS consistency \\ \hline
2 & HowNet & BERTScore, POS consistency \\ \hline
3 & WordNet & USE similarity, POS consistency \\ \hline
\end{tabular}
}
\caption{The three search spaces in our benchmarking. \label{table:search-spaces}} 
\vspace{-3mm}
\end{table}

\begin{table*}[!tbh]
\centering
\scalebox{0.88}{
\begin{tabular}{|p{7.2cm}|c|p{3.5cm}|c|}
\hline
\textbf{Search Algorithm} & \textbf{Deterministic?} & \textbf{Hyperparameters} & \textbf{Num. Queries} \\ \hline
Beam Search \cite{Ebrahimi2017HotFlipWA} & \cmark & $b$ (beam width) & $\mathcal{O}(b * W^2 * T)$ \\ \hline
Greedy [Beam Search with b=1] & \cmark & \centering -- & $\mathcal{O}(W^2 * T)$ \\ \hline
Greedy w. Word Importance Ranking \cite{Gao2018BlackBoxGO,Jin2019TextFooler, pwws-ren-etal-2019-generating} & \cmark & \centering -- & $\mathcal{O}(W * T)$ \\ \hline
Genetic Algorithm \cite{alzantot2018generating} & \xmark & $p$ (population size), $g$ (number of iterations) & $\mathcal{O}(g * p * T)$ \\\hline
Particle Swarm Optimization \cite{pso-zang-etal-2020-word} & \xmark & $p$ (population size), $g$ (number of iterations) & $\mathcal{O}(g * p * W * T)$ \\\hline
\end{tabular}}
\caption{Different search algorithms proposed for NLP attacks. $W$ indicates the number of words in the input. $T$ is the maximum number of transformation options for a given input. \label{table:search-algorithms-categorization}}
\vspace{-5mm}
\end{table*}

\subsection{Heuristic Scoring Function}
\label{heuristic-score}
Search algorithms evaluate potential perturbations before branching out to other solutions. In the case of an untargeted attack against a classifer, the adversary aims to find examples that make the classifier predict the wrong class (label) for $\x'$. Here the assumption is that the ground truth label of $\x'$ is the same as that of the original $\x$.

Naturally, we use a heuristic scoring function $score$ defined as:
\begin{equation}
    score(\x') = 1 - F_y(\x')
\end{equation}
where $F_y(\x)$ is the probability of class $y$ predicted by the model and $y$ is the ground truth output of original text $\x$.

\subsection{Search Algorithms}

We select the following five search algorithms proposed for generating adversarial examples, summarized in Table \ref{table:search-algorithms-categorization}. All search algorithms are limited to modifying each word at most once.

\paragraph{Beam Search}~ 
For given text $\x$, all the possible perturbed texts $\x'$ generated by substituting each word $\x_i$ are scored using the heuristic scoring function, and the top $b$ texts are kept ($b$ is called the "beam width"). Then, the process repeats by further perturbing each of the top $b$ perturbed texts to generate the next set of candidates.

\paragraph{Greedy Search}~ Like beam search, each $\x_i$ are considered for subsitution. We take the best perturbation across all possible perturbations, and repeat until we succeed or run out of possible perturbations. It equals to a beam search with $b$ set to $1$.

\paragraph{Greedy with \importanceRankingNameCap (WIR)}~ 
Words of the given input $\x$ are ranked according to some importance function. Then, in order of descending importance, word $\x_i$ is substituted with $\x'_i$ that maximizes the scoring function until the goal is achieved, or all words have been perturbed. We experiment with four different ways to determine word importance:
\begin{itemize}[noitemsep,topsep=0.1pt]
    \item \textbf{\texttt{UNK}}: Each word's importance is determined by how much the heuristic score changes when the word is substituted with an \texttt{UNK} token \cite{Gao2018BlackBoxGO}. 
    \item \textbf{\texttt{DEL}}: Each word's importance is determined by how much the heuristic score changes when the word is deleted from the original input  \cite{Jin2019TextFooler}.
    \item \textbf{\texttt{PWWS}}: Each word's importance is determined by multiplying the change in score when the word is substituted with an \texttt{UNK} token with the maximum score gained by perturbing the word \cite{pwws-ren-etal-2019-generating}.
    \item \textbf{\texttt{Gradient}}: Similar to how \citet{Wallace2019AllenNLP} visualize saliency of words, each word's importance is determined by calculating the gradient of the loss with respect to the word\footnote{For sub-word tokenization scheme, we take average over all sub-words constituting the word.} and taking its norm.
\end{itemize}
We test an additional scheme, which we call \texttt{RAND}, as an ablation study. Instead of perturbing words in order of their importance, \texttt{RAND} perturbs words in a random order.

\textbf{Genetic Algorithm.} We implement the genetic algorithm of \citet{alzantot2018generating}. At each iteration, each member of the population is perturbed by randomly choosing one word and picking the best $\x'$ gained by perturbing it. Then, crossover occurs between members of the population, with preference given to the more successful members. The algorithm is run for a fixed number of iterations unless it succeeds in the middle. Following \citet{alzantot2018generating}, the population size was 60 and the algorithm was run for at maximum 20 iterations.

\textbf{Particle Swarm Optimization} We implement the particle swarm optimization (PSO) algorithm of \citet{pso-zang-etal-2020-word}. At each iteration, each member of the population is perturbed by first generating all potential $\x'$ obtained by substituting each $\x_i$ and then sampling one $\x'$. Each member is also crossovered with the best perturb text previously found for the member (i.e. local optimum) and the best perturb text found among all members (i.e. global optimum). Following \citet{pso-zang-etal-2020-word}, the population size is set to 60 and the algorithm was run for a maximum of 20 iterations.

Our genetic algorithm and PSO implementations have one small difference from the original implementations. The original implementations contain crossover operations that further perturb the text without considering whether the resulting text meets the defined constraints. In our implementation, we check if the text produced by these subroutines meets our constraints to ensure a consistent search space.

\subsection{Victim Models}
\label{app:victim-models}
We attack \texttt{BERT-base} \cite{devlin2018BERT} and an LSTM fine-tuned on three different datasets:

\begin{itemize}
  \item Yelp polarity reviews \cite{Zhang2015Yelp} (sentiment classification)
  \item Movie Reviews (MR) \cite{pang2015MR} (sentiment classification)
  \item Stanford Natural Language Inference (SNLI) \cite{snli:emnlp2015} (textual entailment).
\end{itemize}

For Yelp and SNLI dataset, we attack $1,000$ samples from the test set, and for MR dataset, we attack $500$ samples. Language of all three datasets is English.

\subsection{Implementation}
We implement all of our attacks using the NLP attack package TextAttack\footnote{TextAttack is available at \url{https://github.com/QData/TextAttack}.} \cite{Morris2020TextAttackAF}. TextAttack provides separate modules for search algorithms, transformations, and constraints, so we can easily compare search algorithms without changing any other part of the attack.

\subsection{Evaluation Metrics}
We use attack success rate ($\frac{\text{\# of successful attacks}}{\text{\# of total attacks}}$) to measure how successful each search algorithm is for attacking a victim model.

To measure the runtime of each algorithm, we use the average number of queries to the victim model as a proxy.

To measure the quality of adversarial examples generated by each algorithm, we use three metrics:
\begin{enumerate}
    \item Average percentage of words perturbed
    \item Universal Sentence Encoder \cite{Cer18USE} similarity between $\x$ and $\x'$
    \item Percent change in perplexities of $\x$ and $\x'$ (using GPT-2 \cite{radford2019language})
\end{enumerate}

\section{Results and Analysis}

\begin{table*}[ht!]
\centering
\scalebox{0.81}{
\begin{tabular}{|c|c|c|cc|cc|cc|}
\toprule
\multirow{2}{*}{\textbf{Model}} & \multirow{2}{*}{\textbf{Dataset}} & \multirow{2}{*}{\textbf{Search Method}} & \multicolumn{2}{|c|}{\textbf{GLOVE Word Embedding}} & \multicolumn{2}{|c|}{\textbf{HowNet}} & \multicolumn{2}{|c|}{\textbf{WordNet}} \\ \cline{4-9} 
 & & & A.S. \% & Avg \# Queries & A.S. \% & Avg \# Queries & A.S. \% & Avg \# Queries \\ \midrule
\multirow{22}{*}{BERT} & \multirow{7}{*}{Yelp} & Greedy (b=1) & $39.5$ & $\commaNum{810}$ & $93.2$ & $\commaNum{3668}$ & $63.2$ & $\commaNum{1480}$ \\
 & & Beam Search (b=4) & $42.0$ & $2857$ & $95.0$ & $\commaNum{10766}$ & $65.9$ & $5033$ \\
 & & Beam Search (b=8) & $42.7$ & $\commaNum{5546}$ & $95.6$ & $\commaNum{19810}$ & $67.3$ & $\commaNum{9674}$ \\
 & & \importanceRankingNameAbbrev (\texttt{UNK}) & $33.2$ & $187$ & $92.3$ & $344$ & $55.3$ & $232$ \\
 & & \importanceRankingNameAbbrev (\texttt{DEL}) & $33.7$ & $189$ & $91.9$ & $364$ & $54.3$ & $238$ \\
 & & \importanceRankingNameAbbrev (\texttt{PWWS}) & $35.3$ & $259$ & $95.1$ & $\commaNum{1300}$ & $58.2$ & $395$ \\
  & & \importanceRankingNameAbbrev (\texttt{Gradient}) & $33.2$ & $55$ & $77.6$ & $189$ & $53.7$ & $94$ \\
 & & \importanceRankingNameAbbrev (\texttt{RAND}) & $29.9$ & $61$ & $72.3$ & $279$ & $53.9$ & $118$ \\
 & & Genetic Algorithm & $37.6$ & $\commaNum{5098}$ & $89.3$ & $\commaNum{11015}$ & $62.1$ & $\commaNum{8257}$ \\
 & & PSO & $\textbf{47.2}$ & $\commaNum{20279}$ & $\textbf{96.6}$ & $\commaNum{62346}$ & $\textbf{74.9}$ & \commaNum{28971} \\
 \cline{2-9}
 & \multirow{7}{*}{MR} & Greedy (b=1) & $20.6$ & $35$ & $78.6$ & $214$ & $59.4$ & $69$ \\
 & & Beam Search (b=4) & $21.4$ & $95$ & $80.6$ & $392$ & $64.6$ & $170$ \\
 & & Beam Search (b=8) & $\textbf{21.8}$ & $175$ & $81.2$ & $632$ & $\textbf{65.8}$ & $303$ \\
 & & \importanceRankingNameAbbrev (\texttt{UNK}) & $17.8$ & $28$ & $53.6$ & $58$ & $55.6$ & $40$ \\
 & & \importanceRankingNameAbbrev (\texttt{DEL}) & $17.0$ & $29$ & $53.6$ & $59$ & $54.0$ & $40$ \\
 & & \importanceRankingNameAbbrev (\texttt{PWWS}) & $21.0$ & $41$ & $73.6$ & $205$ & $58.2$ & $71$ \\
  & & \importanceRankingNameAbbrev (\texttt{Gradient}) & $19.8$ & $14$ & $56.6$ & $46$ & $53.4$ & $24$ \\
 & & \importanceRankingNameAbbrev (\texttt{RAND}) & $17.6$ & $12$ & $48.8$ & $49$ & $53.4$ & $24$ \\
 & & Genetic Algorithm & $\textbf{21.8}$ & $516$ & $80.0$ & $\commaNum{1670}$ & $65.6$ & $\commaNum{1063}$ \\
 & & PSO & $\textbf{21.8}$ & $\commaNum{2413}$ & $\textbf{82.4}$ & $\commaNum{2039}$ & $65.4$ & $\commaNum{2078}$ \\
 \cline{2-9}
 & \multirow{7}{*}{SNLI} & Greedy (b=1) & $19.8$ & $7$ & $87.3$ & $77$ & $49.6$ & $19$ \\
  & & Beam Search (b=4) & $\textbf{20.1}$ & $12$ & $89.2$ & $97$ & $52.0$ & $33$ \\
 & & Beam Search (b=8) & $\textbf{20.1}$ & 18 & $\textbf{89.4}$ & $125$ & $\textbf{52.6}$ & $49$ \\
 & & \importanceRankingNameAbbrev (\texttt{UNK}) & $19.3$ & $22$ & $85.1$ & $47$ & $47.3$ & $30$ \\
 & & \importanceRankingNameAbbrev (\texttt{DEL}) & $18.5$ & $22$ & $84.8$ & $47$ & $46.7$ & $30$ \\
 & & \importanceRankingNameAbbrev (\texttt{PWWS}) & $19.8$ & $26$ & $86.9$ & $116$ & $49.1$ & $42$ \\
 & & \importanceRankingNameAbbrev (\texttt{Gradient}) & $18.8$	& $5$ &  $68.4$	& $25$ & $46.9$ & $10$ \\
 & & \importanceRankingNameAbbrev (\texttt{RAND}) & $18.3$ & $5$ & $82.6$ & $30$ & $46.2$ & $11$ \\
 & & Genetic Algorithm & $20.0$ & $78$ & $89.0$ & $477$ & $52.2$  & $250$ \\
 & & PSO & $\textbf{20.1}$ & $1248$ & $89.1$ & $398$ & $51.9$ & $975$ \\
 \cline{1-9}
 \multirow{22}{*}{LSTM} & \multirow{7}{*}{Yelp} & Greedy (b=1) & $53.0$ & $682$ & $98.2$ & $2611$ &  $80.0$ & $982$ \\
  & & Beam Search (b=4) & $53.2$ & $2313$ & $98.5$ & $7347$ & $81.7$ & $3277$ \\
 & & Beam Search (b=8) & $53.5$ & $4516$ & $98.6$ & $\commaNum{13643}$ & $82.3$ & $\commaNum{6240}$\\
 & & \importanceRankingNameAbbrev (\texttt{UNK}) & $49.3$ & $133$ & $95.2$ & $222$ & $75.8$ & $204$ \\
 & & \importanceRankingNameAbbrev (\texttt{DEL}) & $49.1$ & $181$ & $95.2$ & $230$ & $75.3$ & $205$ \\
 & & \importanceRankingNameAbbrev (\texttt{PWWS}) & $51.2$ & $247$ & $97.3$ & $1212$ & $77.8$ & $361$ \\
 & & \importanceRankingNameAbbrev (\texttt{Gradient}) & $49.3$ & $56$ & $90.0$ & $215$ & $75.3$ & $97$ \\
 & & \importanceRankingNameAbbrev (\texttt{RAND}) & $47.4$ & $57$ & $88.3$ & $217$ & $74.6$ & $98$ \\
 & & Genetic Algorithm & $51.3$ & $5212$ & $98.3$ & $7408$ & $78.5$ & $7245$ \\
 & & PSO & $\textbf{54.9}$ & $\commaNum{17647}$ & $\textbf{98.8}$ & $\commaNum{34659}$  &  $\textbf{84.4}$ & $\commaNum{17145}$ \\
 \cline{2-9}
 & \multirow{7}{*}{MR} & Greedy (b=1) & $38.4$ & $29$ & $87.6$ & $187$ & $74.2$ & $59$ \\
  & & Beam Search (b=4) & $38.6$ & $71$ & $88.6$ & $290$ & $75.6$ & $131$ \\
 & & Beam Search (b=8) & $38.6$ & $127$ & $88.8$ & $427$ & $76.0$ & $222$ \\
 & & \importanceRankingNameAbbrev (\texttt{UNK}) & $35.8$ & $27$ & $81.0$ & $51$ & $72.0$ & $36$ \\
 & & \importanceRankingNameAbbrev (\texttt{DEL}) & $36.2$ & $27$ & $80.2$ & $50$ & $72.2$ & $35$ \\
 & & \importanceRankingNameAbbrev (\texttt{PWWS}) & $37.6$ & $40$ & $86.2$ & $203$ & $73.4$ & $68$ \\
 & & \importanceRankingNameAbbrev (\texttt{Gradient}) & $35.4$ & $10$ & $76.6$ & $36$ & $72.8$ & $18$ \\
 & & \importanceRankingNameAbbrev (\texttt{RAND}) & $34.4$ & $11$  & $68.0$ & $40$ & $71.8$ & $22$ \\
 & & Genetic Algorithm & $\textbf{39.0}$ & $375$ & $88.6$ & $949$ & $76.0$ & $730$ \\
 & & PSO & $\textbf{39.0}$ & $1592$ & $\textbf{89.0}$ & $795$ & $\textbf{76.6}$ & $1179$ \\
 \bottomrule
\end{tabular}}
\caption{Comparison of search methods across three datasets. Models are \texttt{BERT-base} and LSTM fine-tuned for the respective task. ``A.S.\%" represents attack success rate and ``Avg \# Queries" represents the average number of queries made to the model per successful attacked sample.\footnotemark}
\label{table:search-results}
\vspace{-5mm}
\end{table*}
\begin{figure*}[ht!]
\centering
\includegraphics[width=\linewidth]{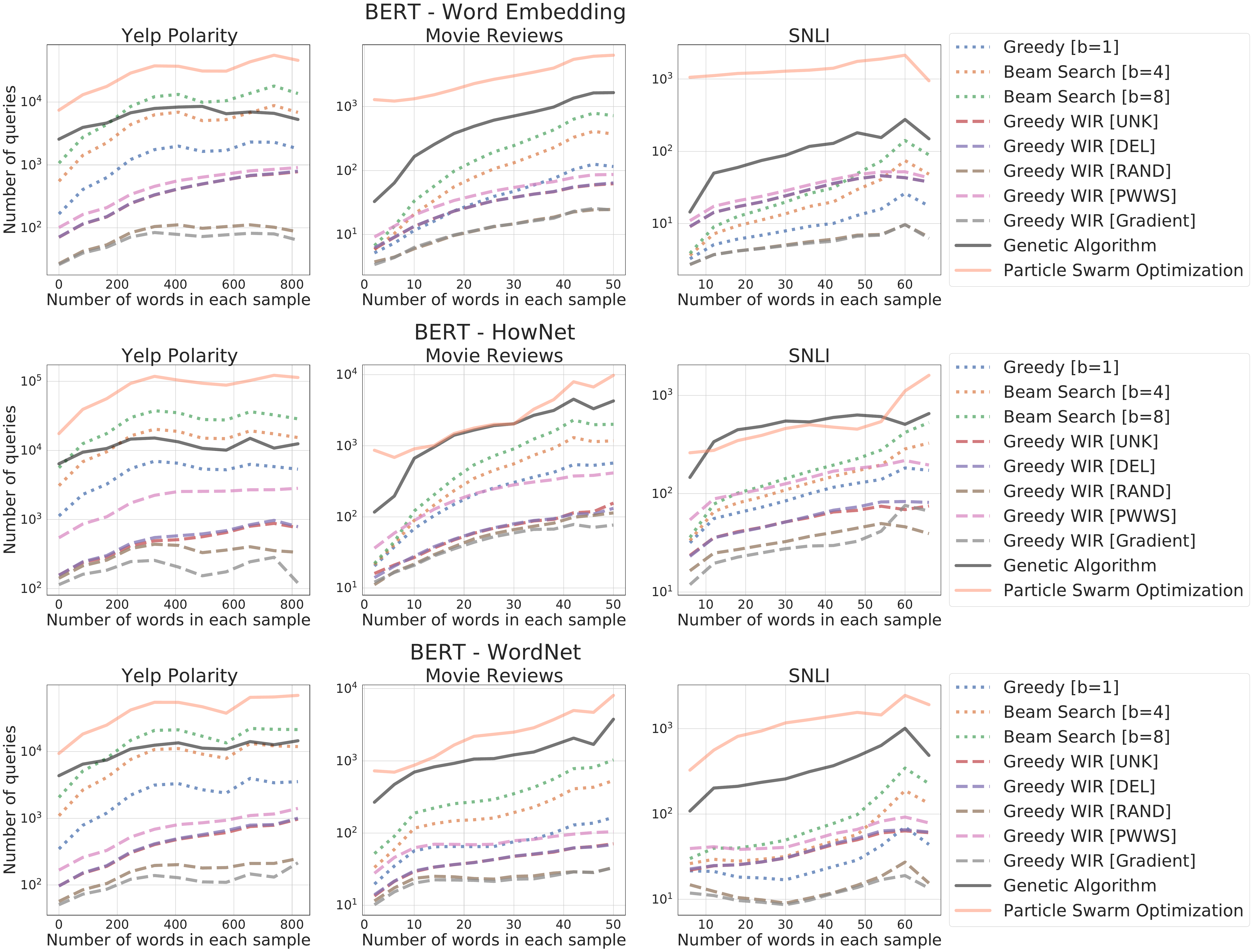}
\caption{Number of queries vs. length of input text. Similar figure for LSTM models are available in the appendix.}
\label{fig:queries-vs-num-words-strict}
\end{figure*}

\begin{figure*}[ht!]
\centering
\includegraphics[width=\linewidth]{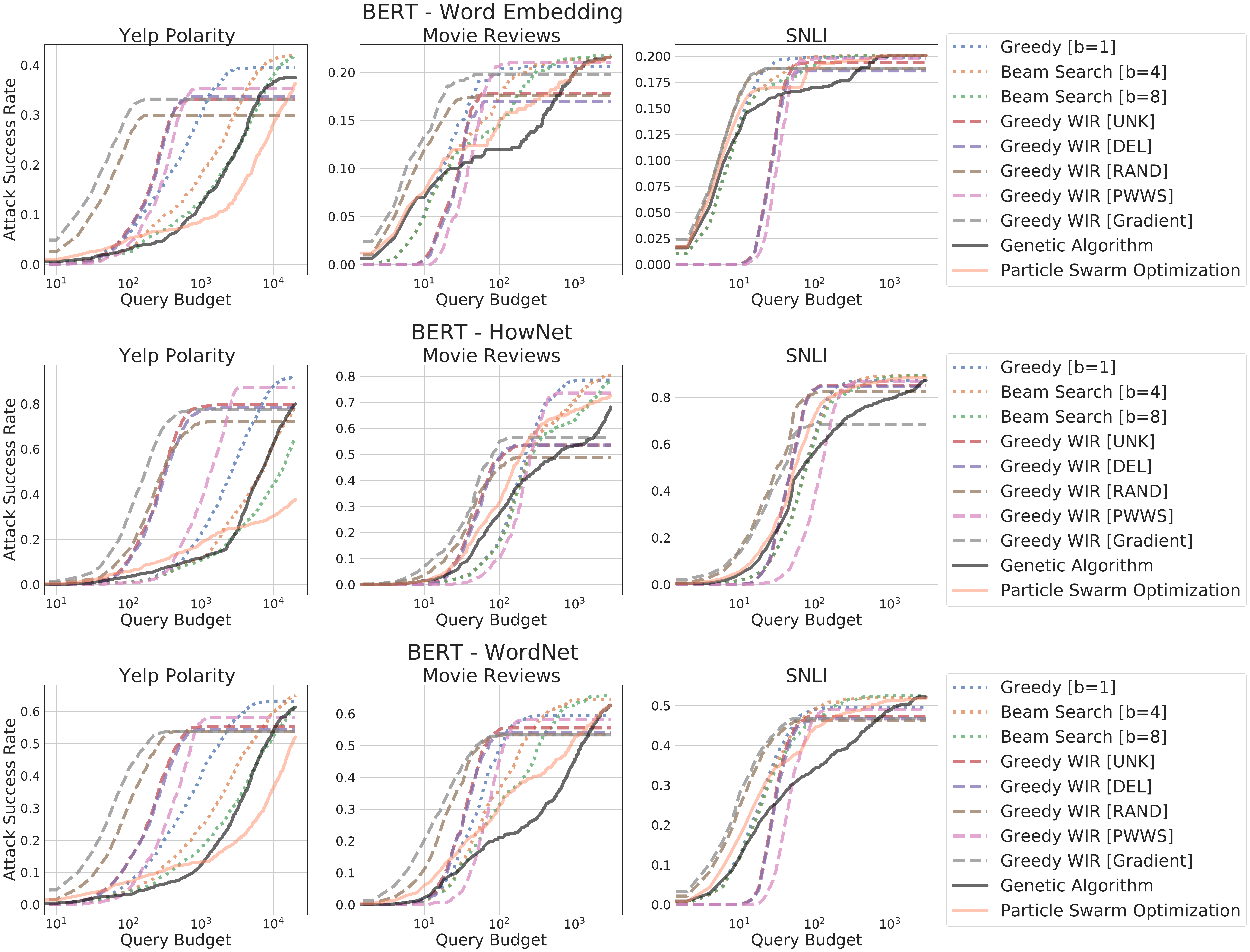}
\caption{Attack success rate by query budget for each search algorithm and dataset. Similar figure for LSTM models are available in the appendix.}
\label{fig:success-by-budget-strict}
\end{figure*}

\subsection{Attack Success Rate Comparison}
Table \ref{table:search-results} shows the results of each attack when each search algorithm is allowed to query the victim model an unlimited number of times. Word importance ranking methods makes far fewer queries than beam or population-based search, while retaining over 60\% of their attack success rate in each case. Beam search (\texttt{b=8}) and PSO are the two most successful search algorithms in every model-dataset combination. However, PSO is more query-intensive. On average, PSO requires 6.3 times\footnote{This is with one outlier (BERT-SNLI with GLOVE word embedding) ignored. If it is included, the number jumps to $10.8$.} more queries than beam search (\texttt{b=8}), but its attack success rate is only on average $1.2\%$ higher than that of beam search (\texttt{b=8}).

\subsection{Runtime Analysis}
Using number of queries to the victim model as proxy for total runtime, Figure \ref{fig:queries-vs-num-words-strict} illustrates how the number of words in the input affects runtime for each algorithm. We can empirically confirm that beam and greedy search algorithms scale quadratically with input length, while \importanceRankingName scales linearly. For shorter datasets, this did not make a significant difference. However, for the longer Yelp dataset, the linear word importance ranking strategies are significantly more query-efficient. These observations match the expected runtimes of the algorithms described in Table \ref{table:search-algorithms-categorization}.

For shorter datasets, genetic and PSO algorithms are significantly more expensive than the other algorithms as the size of population and number of iterations are the dominating factors. Furthermore, PSO is observed to be more expensive than genetic algorithm.

\subsection{Performance under Query Budget}
In a realistic attack scenario, the attacker must conserve the number of queries made to the model. To see which search method was most query-efficient, we calculated the search methods' attack success rates under a range of query budgets. Figure \ref{fig:success-by-budget-strict} shows the attack success rate of each search algorithm as the maximum number of queries permitted to perturb a single sample varies from $0$ to \commaNum{20000} for Yelp dataset and $0$ to \commaNum{3000} for MR and SNLI. 

For both Yelp and MR datasets, the linear (word importance ranking) methods show relatively high success rates within just a few queries, but are eventually surpassed by the slower, quadratic methods (greedy and beam search). The genetic algorithm and PSO lag behind. For SNLI, we see exceptions as the initial queries that linear methods make to determine word importance ranking does not pay off as other algorithms appear more efficient with their queries. This shows that the most effective search method depends on both on the attacker's query budget and the victim model. An attacker with a small query budget may prefer a linear method, but an attacker with a larger query budget may aim to choose a quadratic method to make more queries in exchange for a higher success rate.

Lastly, we can see that both \texttt{Gradient} and \texttt{RAND} ranking methods are initially more successful than \texttt{UNK} and \texttt{DEL} methods, which is due to the overhead involved in calculating word importance ranking for \texttt{UNK} and \texttt{DEL} -- for both methods, each attack makes $W$ queries to determine the importance of each word. Still, \texttt{UNK} and \texttt{DEL} outperform \texttt{RAND} at all but the smallest query budgets, indicating that the order in which words are swapped do matter. Furthermore, in 12 out 15 scenarios, \texttt{UNK} and \texttt{DEL} methods perform as well as or even better than \texttt{Gradient} method, which shows that they are excellent substitutes to the \texttt{Gradient} method for black-box attacks. 

\subsection{Quality of Adversarial Examples}

We selected adversarial examples whose original text $\x$ was successfully attacked by all search algorithms for quality evaluation. Full results of quality evaluation are shown in Table \ref{table:search-autoevaluation} in the appendix. We can see that beam search algorithms consistently perturb the lowest percentage of words. Furthermore, we see that a fewer number of words perturbed generally corresponds with higher average USE similarity between $\x$ and $\x_{adv}$ and a smaller increase in perplexity. This indicates that the beam search algorithms generate higher-quality adversarial examples than other search algorithms. 
\section{Discussion}
\label{s5:discussion}

\subsection{How to Choose A Search Algorithm}
Across all nine scenarios, we can see that choice of search algorithm can have a modest impact on the attack success rate. Query-hungry algorithms such as beam search, genetic algorithm, and PSO perform better than fast WIR methods. Out of the WIR methods, \texttt{PWWS} performs significantly better than \texttt{UNK} and \texttt{DEL} methods. In every case, we see a clear trade-off of performance versus speed. 

With this in mind, one might wonder about what the best way is to choose a suitable search algorithm.
The main factor to consider is the length of the input text. If the input texts are short (e.g. sentence or two), beam search is certainly the appropriate choice: it can achieve a high success rate without sacrificing too much speed. However, when the input text is longer than a few sentences, WIR methods are the most practical choice. If one wishes for the best performance on longer inputs regardless of efficiency, beam search and PSO are the top choices.

\subsection{Effectiveness of \texttt{PWWS} Word Importance Ranking}
Across all tasks, the \texttt{UNK} and \texttt{DEL} methods perform about equivalently, while \texttt{PWWS} performs significantly better than \texttt{UNK} and \texttt{DEL}. In fact, \texttt{PWWS} performs better than greedy search in two cases. However, this gain in performance does come at a cost: \texttt{PWWS} makes far larger number of queries to the victim model to determine the word importance ranking. Out of the 15 experiments, \texttt{PWWS} makes more queries than greedy search in 8 of them. Yet, on average, greedy search outperforms \texttt{PWWS} by $2.5\%$.

Our results question the utility of the \texttt{PWWS} search method. \texttt{PWWS} neither offers the performance that is competitive when compared to greedy search nor the query efficiency that is competitive when compared to \texttt{UNK} or \texttt{DEL}.

\subsection{Effectiveness of Genetic Algorithm}
The genetic algorithm proposed by \citet{alzantot2018generating} uses more queries than the greedy-based beam search (\texttt{b=8}) in 11 of the 15 scenarios, but only achieves a higher attack success rate in 1 scenario. Thus it is generally strictly worse than the simpler beam search (\texttt{b=8}), achieving a lower success rate at a higher cost.

\section{Conclusion}
\vspace{-1mm}

The goal of this paper is not to introduce a new method, but to make empirical analysis towards understanding how search algorithms from recent studies contribute in generating natural language adversarial examples. We evaluated six search algorithms on \texttt{BERT-base} and LSTM models fine-tuned on three datasets. Our results show that when runtime is not a concern, the best-performing methods are beam search and particle swarm optimization. If runtime is of concern, greedy with word importance ranking is the preferable method. We hope that our findings will set a new standard for the reproducibility and evaluation of search algorithms for NLP adversarial examples.

\subsection*{Acknowledgements on Funding:}

This work was partly supported by the National Science Foundation CCF-1900676. Any Opinions, findings and conclusions or recommendations expressed in this material are those of the author(s) and do not necessarily reflect those of the National Science Foundation.

\bibliographystyle{acl_natbib}
\bibliography{search_textattack}

\begin{thebibliography}{30}
\expandafter\ifx\csname natexlab\endcsname\relax\def\natexlab#1{#1}\fi

\bibitem[{Akbik et~al.(2018)Akbik, Blythe, and
  Vollgraf}]{flair-akbik2018coling}
Alan Akbik, Duncan Blythe, and Roland Vollgraf. 2018.
\newblock Contextual string embeddings for sequence labeling.
\newblock In \emph{{COLING} 2018, 27th International Conference on
  Computational Linguistics}, pages 1638--1649.

\bibitem[{Alzantot et~al.(2018)Alzantot, Sharma, Elgohary, Ho, Srivastava, and
  Chang}]{alzantot2018generating}
Moustafa Alzantot, Yash Sharma, Ahmed Elgohary, Bo-Jhang Ho, Mani Srivastava,
  and Kai-Wei Chang. 2018.
\newblock Generating natural language adversarial examples.
\newblock \emph{arXiv preprint arXiv:1804.07998}.

\bibitem[{Bowman et~al.(2015)Bowman, Angeli, Potts, and
  Manning}]{snli:emnlp2015}
Samuel~R. Bowman, Gabor Angeli, Christopher Potts, and Christopher~D Manning.
  2015.
\newblock A large annotated corpus for learning natural language inference.
\newblock In \emph{Proceedings of the 2015 Conference on Empirical Methods in
  Natural Language Processing (EMNLP)}. Association for Computational
  Linguistics.

\bibitem[{Cer et~al.(2018)Cer, Yang, Kong, Hua, Limtiaco, John, Constant,
  Guajardo{-}Cespedes, Yuan, Tar, Sung, Strope, and Kurzweil}]{Cer18USE}
Daniel Cer, Yinfei Yang, Sheng{-}yi Kong, Nan Hua, Nicole Limtiaco, Rhomni~St.
  John, Noah Constant, Mario Guajardo{-}Cespedes, Steve Yuan, Chris Tar,
  Yun{-}Hsuan Sung, Brian Strope, and Ray Kurzweil. 2018.
\newblock \href {http://arxiv.org/abs/1803.11175} {Universal sentence encoder}.
\newblock \emph{CoRR}, abs/1803.11175.

\bibitem[{Chelba et~al.(2013)Chelba, Mikolov, Schuster, Ge, Brants, and
  Koehn}]{google-1-billion-2013}
Ciprian Chelba, Tomas Mikolov, Mike Schuster, Qi~Ge, Thorsten Brants, and
  Phillipp Koehn. 2013.
\newblock \href {http://arxiv.org/abs/1312.3005} {One billion word benchmark
  for measuring progress in statistical language modeling}.
\newblock \emph{CoRR}, abs/1312.3005.

\bibitem[{Devlin et~al.(2018)Devlin, Chang, Lee, and
  Toutanova}]{devlin2018BERT}
Jacob Devlin, Ming{-}Wei Chang, Kenton Lee, and Kristina Toutanova. 2018.
\newblock \href {http://arxiv.org/abs/1810.04805} {{BERT:} pre-training of deep
  bidirectional transformers for language understanding}.
\newblock \emph{CoRR}, abs/1810.04805.

\bibitem[{Dong et~al.(2010)Dong, Dong, and Hao}]{HowNet-2010}
Zhendong Dong, Qiang Dong, and Changling Hao. 2010.
\newblock Hownet and its computation of meaning.
\newblock In \emph{Proceedings of the 23rd International Conference on
  Computational Linguistics: Demonstrations}, COLING ’10, page 53–56, USA.
  Association for Computational Linguistics.

\bibitem[{Ebrahimi et~al.(2017)Ebrahimi, Rao, Lowd, and
  Dou}]{Ebrahimi2017HotFlipWA}
Javid Ebrahimi, Anyi Rao, Daniel Lowd, and Dejing Dou. 2017.
\newblock Hotflip: White-box adversarial examples for text classification.
\newblock In \emph{ACL}.

\bibitem[{Gao et~al.(2018)Gao, Lanchantin, Soffa, and Qi}]{Gao2018BlackBoxGO}
Ji~Gao, Jack Lanchantin, Mary~Lou Soffa, and Yanjun Qi. 2018.
\newblock Black-box generation of adversarial text sequences to evade deep
  learning classifiers.
\newblock \emph{2018 IEEE Security and Privacy Workshops (SPW)}, pages 50--56.

\bibitem[{Goodfellow et~al.(2014)Goodfellow, Shlens, and
  Szegedy}]{goodfellow2014explaining}
Ian~J Goodfellow, Jonathon Shlens, and Christian Szegedy. 2014.
\newblock Explaining and harnessing adversarial examples.
\newblock \emph{arXiv preprint arXiv:1412.6572}.

\bibitem[{Iyyer et~al.(2018)Iyyer, Wieting, Gimpel, and
  Zettlemoyer}]{SCPNs-Iyyer2018-zf}
Mohit Iyyer, John Wieting, Kevin Gimpel, and Luke Zettlemoyer. 2018.
\newblock \href {http://arxiv.org/abs/1804.06059} {Adversarial example
  generation with syntactically controlled paraphrase networks}.

\bibitem[{Jia et~al.(2019)Jia, Raghunathan, G{\"o}ksel, and
  Liang}]{jia2019certified}
Robin Jia, Aditi Raghunathan, Kerem G{\"o}ksel, and Percy Liang. 2019.
\newblock Certified robustness to adversarial word substitutions.
\newblock \emph{arXiv preprint arXiv:1909.00986}.

\bibitem[{Jin et~al.(2019)Jin, Jin, Zhou, and Szolovits}]{Jin2019TextFooler}
Di~Jin, Zhijing Jin, Joey~Tianyi Zhou, and Peter Szolovits. 2019.
\newblock Is bert really robust? natural language attack on text classification
  and entailment.
\newblock \emph{ArXiv}, abs/1907.11932.

\bibitem[{Lei et~al.(2019)Lei, Wu, Chen, Dimakis, Dhillon, and
  Witbrock}]{SubmodularNLPTexas}
Qi~Lei, Lingfei Wu, Pin-Yu Chen, Alex Dimakis, Inderjit~S. Dhillon, and
  Michael~J Witbrock. 2019.
\newblock Discrete adversarial attacks and submodular optimization with
  applications to text classification.
\newblock In \emph{Proceedings of Machine Learning and Systems 2019}, pages
  146--165.

\bibitem[{Miller(1995)}]{wordnet}
George~A. Miller. 1995.
\newblock \href {https://doi.org/10.1145/219717.219748} {Wordnet: A lexical
  database for english}.
\newblock \emph{Commun. ACM}, 38(11):39–41.

\bibitem[{Moon et~al.(2019)Moon, An, and Song}]{Parsimonious-Moon2019-lk}
Seungyong Moon, Gaon An, and Hyun~Oh Song. 2019.
\newblock \href {http://arxiv.org/abs/1905.06635} {Parsimonious {Black-Box}
  adversarial attacks via efficient combinatorial optimization}.

\bibitem[{Morris et~al.(2020{\natexlab{a}})Morris, Lifland, Yoo, and
  Qi}]{Morris2020TextAttackAF}
John Morris, Eli Lifland, Jin~Yong Yoo, and Yanjun Qi. 2020{\natexlab{a}}.
\newblock {TextAttack}: A framework for adversarial attacks in natural language
  processing.
\newblock \emph{ArXiv}, abs/2005.05909.

\bibitem[{Morris et~al.(2020{\natexlab{b}})Morris, Lifland, Lanchantin, Ji, and
  Qi}]{morris2020reevaluating}
John~X. Morris, Eli Lifland, Jack Lanchantin, Yangfeng Ji, and Yanjun Qi.
  2020{\natexlab{b}}.
\newblock \href {http://arxiv.org/abs/2004.14174} {Reevaluating adversarial
  examples in natural language}.

\bibitem[{Mrksic et~al.(2016)Mrksic, S{\'e}aghdha, Thomson, Gasic,
  Rojas-Barahona, hao Su, Vandyke, Wen, and Young}]{Mrksic2016CounterfittingWV}
Nikola Mrksic, Diarmuid~{\'O} S{\'e}aghdha, Blaise Thomson, Milica Gasic,
  Lina~Maria Rojas-Barahona, Pei hao Su, David Vandyke, Tsung-Hsien Wen, and
  Steve~J. Young. 2016.
\newblock Counter-fitting word vectors to linguistic constraints.
\newblock In \emph{HLT-NAACL}.

\bibitem[{Pang and Lee(2005)}]{pang2015MR}
Bo~Pang and Lillian Lee. 2005.
\newblock \href {https://doi.org/10.3115/1219840.1219855} {Seeing stars:
  Exploiting class relationships for sentiment categorization with respect to
  rating scales}.
\newblock In \emph{Proceedings of the 43rd Annual Meeting of the Association
  for Computational Linguistics ({ACL}{'}05)}, pages 115--124, Ann Arbor,
  Michigan. Association for Computational Linguistics.

\bibitem[{Radford et~al.(2019)Radford, Wu, Child, Luan, Amodei, and
  Sutskever}]{radford2019language}
Alec Radford, Jeff Wu, Rewon Child, David Luan, Dario Amodei, and Ilya
  Sutskever. 2019.
\newblock Language models are unsupervised multitask learners.

\bibitem[{Ren et~al.(2019)Ren, Deng, He, and
  Che}]{pwws-ren-etal-2019-generating}
Shuhuai Ren, Yihe Deng, Kun He, and Wanxiang Che. 2019.
\newblock \href {https://doi.org/10.18653/v1/P19-1103} {Generating natural
  language adversarial examples through probability weighted word saliency}.
\newblock In \emph{Proceedings of the 57th Annual Meeting of the Association
  for Computational Linguistics}, pages 1085--1097, Florence, Italy.
  Association for Computational Linguistics.

\bibitem[{Ribeiro et~al.(2018)Ribeiro, Singh, and
  Guestrin}]{SEARS-Ribeiro2018-ue}
Marco~Tulio Ribeiro, Sameer Singh, and Carlos Guestrin. 2018.
\newblock Semantically equivalent adversarial rules for debugging {NLP} models.
\newblock pages 856--865.

\bibitem[{Szegedy et~al.(2013)Szegedy, Zaremba, Sutskever, Bruna, Erhan,
  Goodfellow, and Fergus}]{szegedy2013intriguing}
Christian Szegedy, Wojciech Zaremba, Ilya Sutskever, Joan Bruna, Dumitru Erhan,
  Ian Goodfellow, and Rob Fergus. 2013.
\newblock Intriguing properties of neural networks.
\newblock \emph{arXiv preprint arXiv:1312.6199}.

\bibitem[{Wallace et~al.(2019)Wallace, Tuyls, Wang, Subramanian, Gardner, and
  Singh}]{Wallace2019AllenNLP}
Eric Wallace, Jens Tuyls, Junlin Wang, Sanjay Subramanian, Matt Gardner, and
  Sameer Singh. 2019.
\newblock {AllenNLP Interpret}: A framework for explaining predictions of {NLP}
  models.
\newblock In \emph{Empirical Methods in Natural Language Processing}.

\bibitem[{Zang et~al.(2020)Zang, Qi, Yang, Liu, Zhang, Liu, and
  Sun}]{pso-zang-etal-2020-word}
Yuan Zang, Fanchao Qi, Chenghao Yang, Zhiyuan Liu, Meng Zhang, Qun Liu, and
  Maosong Sun. 2020.
\newblock \href {https://www.aclweb.org/anthology/2020.acl-main.540}
  {Word-level textual adversarial attacking as combinatorial optimization}.
\newblock In \emph{Proceedings of the 58th Annual Meeting of the Association
  for Computational Linguistics}, pages 6066--6080, Online. Association for
  Computational Linguistics.

\bibitem[{Zhang et~al.(2019{\natexlab{a}})Zhang, Zhou, Miao, and
  Li}]{zhang2019mha}
Huangzhao Zhang, Hao Zhou, Ning Miao, and Lei Li. 2019{\natexlab{a}}.
\newblock \href {https://doi.org/10.18653/v1/P19-1559} {Generating fluent
  adversarial examples for natural languages}.
\newblock In \emph{Proceedings of the 57th Annual Meeting of the Association
  for Computational Linguistics}, pages 5564--5569, Florence, Italy.
  Association for Computational Linguistics.

\bibitem[{Zhang* et~al.(2020)Zhang*, Kishore*, Wu*, Weinberger, and
  Artzi}]{bert-score}
Tianyi Zhang*, Varsha Kishore*, Felix Wu*, Kilian~Q. Weinberger, and Yoav
  Artzi. 2020.
\newblock \href {https://openreview.net/forum?id=SkeHuCVFDr} {Bertscore:
  Evaluating text generation with bert}.
\newblock In \emph{International Conference on Learning Representations}.

\bibitem[{Zhang et~al.(2019{\natexlab{b}})Zhang, Sheng, and
  Alhazmi}]{Zhang19Survey}
Wei~Emma Zhang, Quan~Z. Sheng, and Ahoud Abdulrahmn~F. Alhazmi.
  2019{\natexlab{b}}.
\newblock \href {http://arxiv.org/abs/1901.06796} {Generating textual
  adversarial examples for deep learning models: {A} survey}.
\newblock \emph{CoRR}, abs/1901.06796.

\bibitem[{Zhang et~al.(2015)Zhang, Zhao, and LeCun}]{Zhang2015Yelp}
Xiang Zhang, Junbo Zhao, and Yann LeCun. 2015.
\newblock Character-level convolutional networks for text classification.
\newblock In C.~Cortes, N.~D. Lawrence, D.~D. Lee, M.~Sugiyama, and R.~Garnett,
  editors, \emph{Advances in Neural Information Processing Systems 28}, pages
  649--657. Curran Associates, Inc.

\end{thebibliography}

\cleardoublepage

\appendix

\section{Appendix on Results}
\subsection{Search Space Choices}
\label{exp:space}

Here, we provide more detail about our search space choices. 

\subsubsection{Transformations}
We consider three different ways of generating substitute words:
\begin{enumerate}
    \item Counter-fitted GLOVE word embedding \cite{Mrksic2016CounterfittingWV}: For a given word, we take its top $N$ nearest neighbors in the embedding space as its synonyms. \footnote{We choose counter-fitted embeddings because they encode synonym/antonym representations better than vanilla GLoVe embeddings \cite{Mrksic2016CounterfittingWV}.}
    \item HowNet \cite{HowNet-2010}: HowNet is a knowledge base of sememes in both Chinese and English.
    \item WordNet \cite{wordnet}: WordNet is a lexical database that contains knowledge about and relationships between English words, including synonyms.
\end{enumerate}

\subsubsection{Constraints}
To preserve grammaticality, we require that the two words being swapped have the same part-of-speech (POS). This is determined by a part-of-speech tagger provided by Flair \cite{flair-akbik2018coling}, an open-source NLP library.

To preserve semantics, we consider three different constraints:
\begin{enumerate}
    \item Minimum cosine similarity of word embeddings: For the word embedding transformation, we require the cosine similarity of the embeddings of the two words meet a minimum threshold.
    
    \item Minimum BERTScore \cite{bert-score}: We require that the F1 BERTScore between $x$ and $x'$ meet some minimum threshold value. 
    
    \item Universal Sentence Encoder \cite{Cer18USE}: We require that the angular similarity between the sentence embeddings of $x$ and $x'$ meet some minimum threshold. 
\end{enumerate}
For word embedding similarity, BERTScore, and USE similarity, we need to set the minimum threshold value. We set all three values to be $0.9$ based on the observation reported by \citet{morris2020reevaluating} that high threshold values encourages strong semantic similarity. We do not apply word embedding similarity constraint for HowNet and WordNet transformations because it is not guaranteed that we can map the substitute words generated from the two sources to a word embedding space. We can also assume that the substitute words are semantically similar to the original words since they originate from a curated knowledge base.

Lastly, for all attacks carried out, we do not allow perturbing a word that has already been perturbed and we do not perturbed pre-defined stop words.

\label{app:datasets}

\subsubsection{Datasets}
\begin{figure*}[tbh]
\centering
\includegraphics[width=\linewidth]{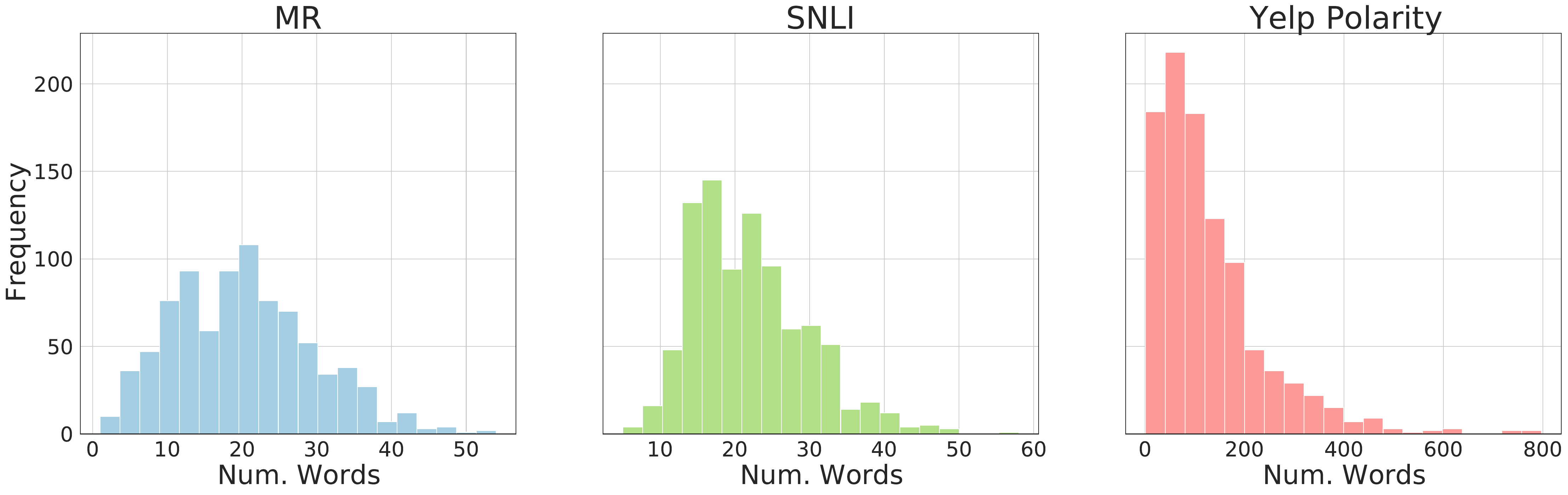}
\caption{Histogram of words per dataset. Yelp inputs are generally much longer than inputs from MR or SNLI.}
\label{fig:dataset-wordcount-hist}
\end{figure*}

We compare search algorithms on three datasets: the Movie Review and Yelp Polarity sentiment classification datasets and the SNLI entailment dataset. Figure \ref{fig:dataset-wordcount-hist} shows a histogram of the number of words in inputs from each dataset. We can see that inputs from Yelp are generally much longer than inputs from MR or SNLI.

\algrenewcommand\algorithmicrequire{\textbf{Input:}}
\algrenewcommand\algorithmicensure{\textbf{Output:}}
\newcommand*\Let[2]{\State #1 $\gets$ #2}

\subsection{Pseudocode for Search Algorithms}
\label{app:pseudocode}
Before presenting the pseudocode of each search algorithm, we define a subroutine called $perturb$ that takes text $x$ and index $i$ to produce set of perturbation $x'$ that satisfies the constraints. More specifically, $perturb$ is defined as following:
\begin{multline*}
    perturb(x, i) = \\
    \{T(x,i)\; | \; C_j(T(x,i)) \; \forall j \in \{1,...,m\}\}
\end{multline*}
where $T(\x, i)$ represents the transformation method that swaps the $i^{th}$ word $x_i$ with its synonym to produce perturbed text$\x'$. $C_1, \dots, C_m$ are constraints represented as Boolean functions. $C_i(x) = True$ means that text $x$ satisfies constraint $C_i$. 

Also, $score(x)$ is the heuristic scoring function that was defined in the section \ref{heuristic-score}.
\newline

\begin{algorithm}  
\caption{Beam Search with beam width $b$}  
\begin{algorithmic}[5]  
    \Require{Original text $x=(x_1, x_2, ... x_n)$}
    \Ensure{Adversarial text $x_{adv}$ if found}
    \Let{$best$}{$\{x\}$}
    \While{$best == \emptyset$}
        \Let{$X_{cand}$}{$\emptyset$}
        \ForAll{$x_b \in best$}
            \ForAll{$i \in \{1,\dots,n\}$}
                \Let{$X_{cand}$}{$X_{cand} \cup perturb(x_b, i)$}
            \EndFor
        \EndFor
        \If{$X_{cand} \neq \emptyset$}
            \Let{$x^*$}{$\argmax_{x'\in X_{cand}}{score(x')}$}
            \If{$x^*$ fools the model}
                \State \Return{$x^*$ as $x_{adv}$}
            \Else
                \Let{$best$}{$\{\text{top } b$ elements of $X_{cand}\}$} \\
                \Comment{elements are ranked by their $score$}
            \EndIf
        \Else
            \State{End search}
        \EndIf
    \EndWhile
\end{algorithmic}  
\end{algorithm}

\begin{algorithm}  
\caption{Greedy Search}  
\begin{algorithmic}[5]  
    \Require{Original text $x=(x_1, x_2, ... x_n)$}
    \Ensure{Adversarial text $x_{adv}$ if found}
    \Let{$x^*$}{$x$}
    \While{$x_{adv}$ not found}
        \Let{$X_{cand}$}{$\emptyset$}
        \ForAll{$i \in \{1,\dots,n\}$}
            \Let{$X_{cand}$}{$X_{cand} \cup perturb(x^*, i)$}
        \EndFor

        \If{$X_{cand} \neq \emptyset$}
            \Let{$x^*$}{$\argmax_{x'\in X_{cand}}{score(x')}$}
            \If{$x^*$ fools the model}
                \State \Return{$x^*$ as $x_{adv}$}
            \EndIf
        \Else
            \State{End search}
        \EndIf
    \EndWhile
\end{algorithmic}  
\end{algorithm}

Greedy search with word importance ranking requires subroutine for determining the importance of each word in text $x$. We leave the details of the importance functions to be found in individual papers that have proposed them, including \citet{Gao2018BlackBoxGO}, \citet{Jin2019TextFooler},  \citet{pwws-ren-etal-2019-generating}.
\begin{algorithm}  
\caption{Greedy Search with Word Importance Ranking}  
\begin{algorithmic}[5]  
    \Require{Original text $x=(x_1, x_2, ... x_n)$}
    \Ensure{Adversarial text $x_{adv}$ if found}
    \Let{$R$}{ranking $r_1,\dots,r_n$ of words $x_1,\dots,x_n$ by their importance}
    \Let{$x^*$}{$x$}
    \For{$i = r_1, r_2,\dots, r_n$ in $R$ }
        \Let{$X_{cand}$}{$perturb(x^*, i)$}
        \If{$X_{cand} \neq \emptyset$}
            \Let{$x^*$}{$\argmax_{x'\in X_{cand}}{score(x')}$}
            \If{$x^*$ fools the model}
                \State \Return{$x^*$ as $x_{adv}$}
            \EndIf
        \Else
            \State{End search}
        \EndIf
    \EndFor

\end{algorithmic}  
\end{algorithm}

In genetic algorithm, each population member represents a distinct text produced via $perturb$ and $crossover$ operations. Genetic algorithm has a subroutine called $sample$ that takes in population member $p$ and randomly samples a word to transform with probabilities proportional to the number of synonyms a word has. Also, we modified the $crossover$ subroutine proposed by \citet{alzantot2018generating} to check if child produced by $crossover$ operation passes constraints. If the child fails any of the constraints, we retry the crossover for at max 20 times. If that also fails to produce a child that passes constraints, we randomly choose one its parents to be the child with equal probability.

\begin{algorithm}  
\caption{Genetic Algorithm (with population size $K$ and generation $G$)}  
\begin{algorithmic}[5]  
    \Require{Original text $x=(x_1, x_2, ... x_n)$}
    \Ensure{Adversarial text $x_{adv}$ if found}
    \For{$k = 1,\dots,K$}
        \Let{$i$}{$sample(x)$}
        \Let{$X_{cand}$}{$perturb(x, i)$}
        \Let{$P^0_k$}{$\argmax_{x'\in X_{cand}}{score(x')}$}
    \EndFor
    \For{$g = 1,\dots,G$ generations}
        \For{$k = 1,\dots,K$}
            \Let{$S^{g-1}_k$}{$score(P^{g-1}_k)$}
        \EndFor
        \Let{$x^*$}{$P^{g-1}_{\argmax_{j}{S^{g-1}_j}}$}
        \If{$x^*$ fools the model}
            \State \Return{$x^*$ as $x_{adv}$}
        \Else
            \Let{$P^g_1$}{$x^*$}
            \State{$p=Normalize(S^{g-1})$}
            \For{$k = 2,\dots, K$}
                \State{$par_1 \sim P^{g-1}$ with prob $p$}
                \State{$par_2 \sim P^{g-1}$ with prob $p$}
                \Let{$child$}{$crossover(par_1, par_2)$}
                \Let{$i$}{$sample(child)$}
                \Let{$X_{cand}$}{$perturb(child, i)$}
                \Let{$P^g_k$}{$\argmax_{x'\in X_{cand}}{score(x')}$}
            \EndFor
        \EndIf
    \EndFor

\end{algorithmic}  
\end{algorithm}

Lastly, we leave out the pseudocode for PSO due to its complexity. More detail can be found in \citet{pso-zang-etal-2020-word}.

\subsection{Analysis of Attacks against LSTM Models}
\label{app:lstm-figures}

\begin{figure*}[ht!]
\centering
\includegraphics[width=\linewidth]{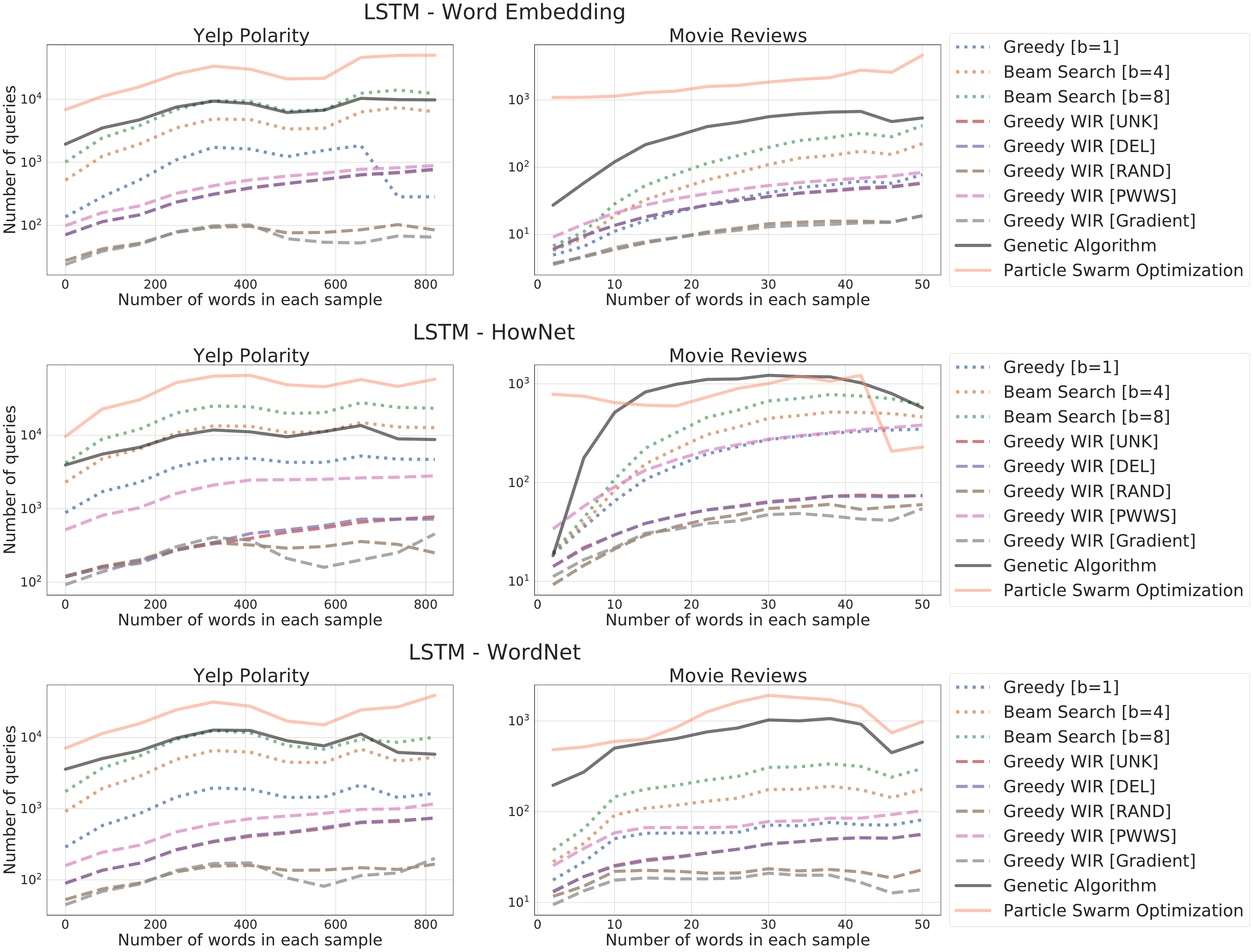}
\caption{Number of queries vs. length of input text.}
\label{fig:lstm-queries-vs-num-words-strict}
\end{figure*}

\begin{figure*}[ht!]
\centering
\includegraphics[width=\linewidth]{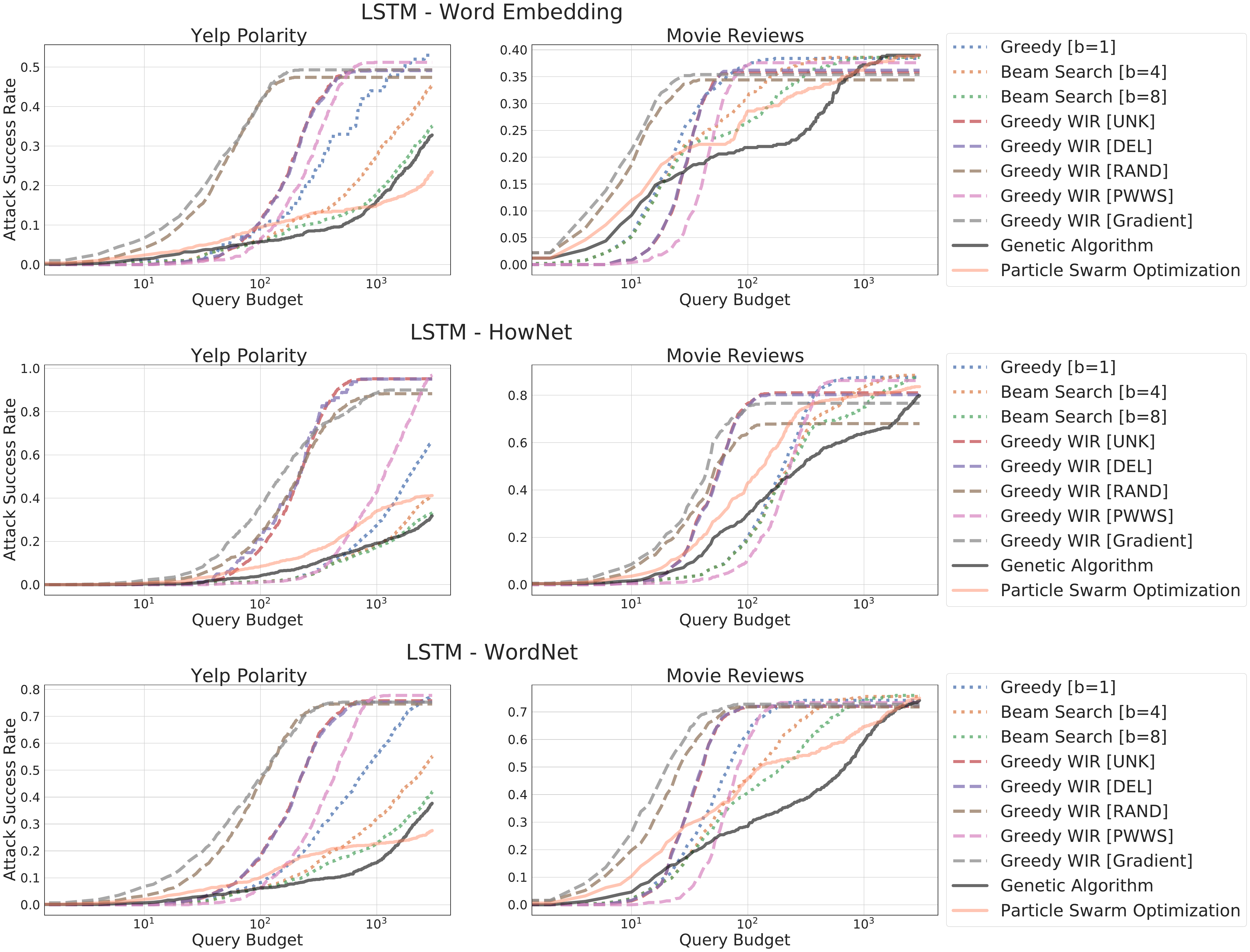}
\caption{Attack success rate by query budget for each search algorithm and dataset.}
\label{fig:lstm-success-by-budget-strict}
\end{figure*}

Figures \ref{fig:lstm-queries-vs-num-words-strict} shows how the number of words in the input affects runtime for each algorithm against LSTM models. Figure \ref{fig:lstm-success-by-budget-strict} shows the attack success rate of each search algorithm as the maximum number of queries permitted to perturb a single sample varies from $0$ to \commaNum{20000} for Yelp dataset and $0$ to \commaNum{3000} for MR and SNLI. 

\subsection{Evaluation of Adversarial Examples}
\label{app:eval-ae-table}
Table \ref{table:search-autoevaluation} shows the average percentage of words perturbed, average Universal Sentence Encoder similarity score, and average percent change in perplexity for all experiments. 
\begin{table*}[ht!]
\centering
\scalebox{0.55}{
\begin{tabular}{|c|c|c|ccc|ccc|ccc|}
\toprule
\multirow{2}{*}{Model} & \multirow{2}{*}{Dataset} & \multirow{2}{*}{Search Method} & \multicolumn{3}{|c|}{GLOVE Word Embedding} & \multicolumn{3}{|c|}{HowNet} & \multicolumn{3}{|c|}{WordNet} \\ \cline{4-12} 
 & & & Avg P.W. \% & Avg USE Sim  & $\Delta\%$ Perplexity & Avg P.W. \% & Avg USE Sim  & $\Delta\%$ Perplexity & Avg P.W. \% & Avg USE Sim  & $\Delta\%$ Perplexity \\ \midrule
\multirow{22}{*}{BERT} & \multirow{7}{*}{Yelp} & Greedy (b=1) &3.41 & 0.948 & 21.5 & 2.52 & 0.945 & 22.8 & 4.76 & 0.943 & 49.9\\
& & Beam Search (b=4) &3.26 & 0.949 & 20.7 & 2.45 & 0.946 & 22.0 & 4.49 & 0.944 & 46.7\\
& & Beam Search (b=8) & \textbf{3.20} & \textbf{0.950} & \textbf{20.1} & \textbf{2.42} & \textbf{0.947} & \textbf{21.4} & \textbf{4.46} & \textbf{0.945} & \textbf{46.4}\\
& & \importanceRankingNameAbbrev (\texttt{UNK}) &6.48 & 0.930 & 43.5 & 4.73 & 0.922 & 42.3 & 9.02 & 0.924 & 92.1\\
& & \importanceRankingNameAbbrev (\texttt{DEL}) &6.85 & 0.928 & 47.2 & 5.10 & 0.919 & 46.4 & 9.38 & 0.923 & 98.8\\
& & \importanceRankingNameAbbrev (\texttt{PWWS}) &4.36 & 0.942 & 27.3 & 3.11 & 0.94 & 28.1 & 6.10 & 0.937 & 66.1\\
& & \importanceRankingNameAbbrev (\texttt{Gradient}) &6.16 & 0.933 & 37.8 & 5.58 & 0.913 & 44.5 & 9.10 & 0.925 & 86.4\\
& & \importanceRankingNameAbbrev (\texttt{RAND}) &8.18 & 0.920 & 59.1 & 7.46 & 0.898 & 74.6 & 11.16 & 0.914 & 124.8\\
& & Genetic Algorithm &5.06 & 0.936 & 33.9 & 4.21 & 0.928 & 42.7 & 6.70 & 0.932 & 77.3\\
& & PSO &6.61 & 0.929 & 47.3 & 6.08 & 0.913 & 62.3 & 9.67 & 0.922 & 111.0\\
 \cline{2-12}
 & \multirow{7}{*}{MR} & Greedy (b=1) &7.25 & 0.900 & 31.8 & 6.14 & 0.887 & 36.5 & 10.26 & 0.864 & 102.8\\
& & Beam Search (b=4) & \textbf{7.22} & \textbf{0.901} & \textbf{31.4} & \textbf{6.10} & \textbf{0.887} & \textbf{36.1} & 10.10 & \textbf{0.866} & \textbf{97.9}\\
& & Beam Search (b=8) & \textbf{7.22} & \textbf{0.901} & \textbf{31.4} & \textbf{6.10} & \textbf{0.887} & \textbf{36.1} & \textbf{10.05} & \textbf{0.866} & 101.6\\
& & \importanceRankingNameAbbrev (\texttt{UNK}) &9.42 & 0.884 & 42.3 & 7.77 & 0.866 & 48.0 & 14.14 & 0.845 & 141.2\\
& & \importanceRankingNameAbbrev (\texttt{DEL}) &9.62 & 0.882 & 46.4 & 7.69 & 0.865 & 46.1 & 14.60 & 0.840 & 146.4\\
& & \importanceRankingNameAbbrev (\texttt{PWWS}) &7.36 & 0.898 & 33.8 & 6.22 & 0.884 & 37.6 & 10.80 & 0.865 & 111.1\\
& & \importanceRankingNameAbbrev (\texttt{Gradient}) &8.61 & 0.892 & 38.1 & 8.25 & 0.862 & 40.8 & 14.58 & 0.844 & 123.2\\
& & \importanceRankingNameAbbrev (\texttt{RAND}) &10.1 & 0.881 & 51.4 & 9.93 & 0.846 & 69.5 & 17.28 & 0.827 & 149.4\\
& & Genetic Algorithm &8.18 & 0.895 & 35.8 & 6.41 & 0.885 & 37.8 & 12.30 & 0.854 & 124.5\\
& & PSO &8.71 & 0.894 & 39.0 & 6.46 & 0.884 & 38.7 & 16.08 & 0.839 & 187.8\\
 \cline{2-12}
 & \multirow{7}{*}{SNLI} & Greedy (b=1) &5.59 & 0.915 & 37.8 & 5.02 & 0.889 & 31.7 & 6.53 & 0.903 & 55.9\\
& & Beam Search (b=4) & \textbf{5.59} & \textbf{0.916} & \textbf{37.8} & \textbf{5.02} & \textbf{0.889} & \textbf{31.6} & \textbf{6.50} & \textbf{0.903} & \textbf{55.7}\\
& & Beam Search (b=8) & 5.59 & 0.916 & 37.8 & 5.02 & 0.889 & 31.6 & 6.50 & 0.903 & 55.9\\
& & \importanceRankingNameAbbrev (\texttt{UNK}) &6.56 & 0.911 & 42.8 & 5.65 & 0.887 & 33.4 & 8.03 & 0.899 & 65.5\\
& & \importanceRankingNameAbbrev (\texttt{DEL}) &6.77 & 0.91 & 44.0 & 5.81 & 0.887 & 34.2 & 8.22 & 0.898 & 67.6\\
& & \importanceRankingNameAbbrev (\texttt{PWWS}) &5.63 & 0.915 & 37.8 & 5.05 & 0.89 & 30.5 & 6.59 & 0.906 & 54.5\\
& & \importanceRankingNameAbbrev (\texttt{Gradient}) &6.57 & 0.911 & 41.6 & 5.9 & 0.881 & 37.7 & 8.06 & 0.899 & 65.1\\
& & \importanceRankingNameAbbrev (\texttt{RAND}) &7.06 & 0.909 & 47.7 & 6.19 & 0.884 & 42.9 & 8.65 & 0.895 & 74.6\\
& & Genetic Algorithm &5.71 & 0.915 & 38.5 & 5.14 & 0.888 & 32.7 & 6.73 & 0.902 & 58.3\\
& & PSO &5.76 & 0.915 & 38.6 & 5.14 & 0.888 & 32.5 & 6.94 & 0.902 & 58.5\\
  \cline{1-12}
 \multirow{22}{*}{LSTM} & \multirow{7}{*}{Yelp} & Greedy (b=1) & 4.04 & 0.943 & 28.9 & 2.47 & 0.948 & 23.9 & 4.58 & 0.946 & 52.1 \\
 & & Beam Search (b=4) & 4.01 & 0.942 & 28.9 & 2.47 & 0.949 & 23.7 & 4.53 & 0.946 & 51.9\\
 & & Beam Search (b=8) & \textbf{4.01} & \textbf{0.943} & \textbf{28.7} & \textbf{2.44} & \textbf{0.949} & \textbf{23.0} & \textbf{4.51} & \textbf{0.946} & \textbf{51.3}\\
 & & \importanceRankingNameAbbrev (\texttt{UNK}) & 5.83 & 0.933 & 42.4 & 3.51 & 0.935 & 34.4 & 7.22 & 0.935 & 75.6\\
& & \importanceRankingNameAbbrev (\texttt{DEL}) & 5.86 & 0.932 & 41.1 & 3.61 & 0.936 & 33.3 & 7.22 & 0.935 & 75.4\\
& & \importanceRankingNameAbbrev (\texttt{PWWS}) & 4.57 & 0.940 & 32.6 & 2.58 & 0.947 & 23.9 & 5.14 & 0.944 & 57.0\\
& & \importanceRankingNameAbbrev (\texttt{Gradient}) & 7.05 & 0.926 & 52.2 & 5.25 & 0.916 & 50.7 & 8.42 & 0.929 & 87.9\\
 & & \importanceRankingNameAbbrev (\texttt{RAND}) & 7.28 & 0.925 & 53.6 & 6.33 & 0.906 & 69.5 & 9.40 & 0.925 & 102.4\\
& & Genetic Algorithm & 5.94 & 0.933 & 42.8 & 3.73 & 0.930 & 41.9 & 6.37 & 0.936 & 80.9\\
& & PSO & 6.70 & 0.929 & 47.3 & 5.03 & 0.924 & 58.7 & 7.98 & 0.93 & 95.2\\
 \cline{2-12}
& \multirow{7}{*}{MR} & Greedy (b=1) & \textbf{7.19} & \textbf{0.899} & \textbf{33.5} & \textbf{5.96} & 0.884 & 37.2 & 10.21 & 0.871 & 100.6\\
& & Beam Search (b=4) & 7.19 & 0.899 & 33.7 & 5.96 & 0.884 & 37.6 & 10.03 & 0.871 & 98.7\\
& & Beam Search (b=8) & 7.19 & 0.899 & 34.0 & 5.96 & 0.884 & 37.6 & \textbf{10.00} & \textbf{0.871} & 97.4\\
& & \importanceRankingNameAbbrev (\texttt{UNK}) & 8.99 & 0.889 & 41.7 & 7.22 & 0.874 & 42.9 & 12.99 & 0.856 & 104.5\\
& & \importanceRankingNameAbbrev (\texttt{DEL}) & 9.17 & 0.889 & 44.5 & 7.21 & 0.874 & 42.2 & 13.03 & 0.856 & 107.5\\
& & \importanceRankingNameAbbrev (\texttt{PWWS}) & 7.45 & 0.898 & 33.7 & 6.01 & 0.884 & \textbf{37.1} & 10.50 & 0.871 & \textbf{87.9}\\
& & \importanceRankingNameAbbrev (\texttt{Gradient}) & 8.73 & 0.892 & 41.4 & 7.33 & 0.870 & 40.8 & 13.12 & 0.859 & 104.1\\
& & \importanceRankingNameAbbrev (\texttt{RAND}) & 10.60 & 0.880 & 54.0 & 9.31 & 0.853 & 57.7 & 16.05 & 0.842 & 148.2\\
& & Genetic Algorithm & 8.02 & 0.896 & 36.4 & 6.36 & 0.881 & 39.5 & 11.98 & 0.860 & 120.2\\
 & & PSO & 8.41 & 0.893 & 40.2 & 6.32 & 0.882 & 40.8 & 13.90 & 0.854 & 130.0 \\

 \bottomrule
\end{tabular}}
\caption{Quality evaluation of the adversarial examples produced by each search algorithm. "Avg P.W. \%" means average percentage of words perturbed, "Avg USE Sim" means average USE angular similarity, and "$\Delta \%$ Perplexity" means percent change in perplexities.}
\label{table:search-autoevaluation}
\end{table*}
\section{Future Work}

\paragraph{Submodularity of transformer models.}~ As mentioned in Section \ref{s5:discussion}, our findings indicate that the NLP attack problem may be approximately submodular when dealing with transformer models. In the image space, attacks designed to take advantage of submodularity have achieved high query efficiency \cite{Parsimonious-Moon2019-lk}. With the exception of \cite{SubmodularNLPTexas}, attacks in NLP are yet to take advantage of this submodular property.

\paragraph{Transformations beyond word-level.}~ Most proposed adversarial attacks in NLP focus on making substitutions at the word level or the character level. A few works have considered replacing phrases \cite{SEARS-Ribeiro2018-ue} as well as paraphrasing full sentences \cite{SubmodularNLPTexas,SCPNs-Iyyer2018-zf}. However, neither of these scenarios has been studied extensively. Future work in NLP adversarial examples would benefit from further exploration of phrase and sentence-level transformations.

\paragraph{Motivations for generating NLP adversarial examples.}~ One purpose of generating adversarial examples for NLP systems is to improve the systems. Much work has focused on improvements in intrinsic evaluation metrics like achieving higher attack success rate via an improved search method. To advance the field, future researchers might focus more on using adversarial examples in NLP to build better NLP systems.

\end{document}